# Dynamic Attention Mechanism in Spatiotemporal Memory Networks for Object Tracking

Meng Zhou, Jiadong Xie, and Mingsheng Xu

*Abstract*—Mainstream visual object tracking frameworks predominantly rely on template matching paradigms. Their performance heavily depends on the quality of template features, which becomes increasingly challenging to maintain in complex scenarios involving target deformation, occlusion, and background clutter. While existing spatiotemporal memory-based trackers emphasize memory capacity expansion, they lack effective mechanisms for dynamic feature selection and adaptive fusion. To address this gap, we propose a Dynamic Attention Mechanism in Spatiotemporal Memory Network (DASTM) with two key innovations: 1) A differentiable dynamic attention mechanism that adaptively adjusts channel-spatial attention weights by analyzing spatiotemporal correlations between the templates and memory features; 2) A lightweight gating network that autonomously allocates computational resources based on target motion states, prioritizing high-discriminability features in challenging scenarios. Extensive evaluations on OTB-2015, VOT 2018, LaSOT, and GOT-10K benchmarks demonstrate our DASTM's superiority, achieving state-of-the-art performance in success rate, robustness, and real-time efficiency, thereby offering a novel solution for real-time tracking in complex environments.

## I. INTRODUCTION

With the rapid advancement of deep learning technologies and continuous improvements in hardware capabilities, the field of computer vision has achieved groundbreaking progress. Visual object tracking[1], as a significant branch of computer vision, has demonstrated its application value across various domains. However, with the diversification of application scenarios, the complexity of tracking has also increased. Existing tracking methods still face challenges such as occlusion, blurring, rapid deformation, and interference from similar targets in complex real-world environments. Therefore, visual object tracking remains a highly valuable research area, necessitating more precise and robust trackers to address these complex scenarios.

In recent years, Siamese network-based methods[2],[3],[4],[5],[6],[7]have provided a novel framework for object tracking tasks, simplifying the process of matching target locations. However, these methods often rely on fixed templates, which may lead to tracking failures when encountering complex situations such as occlusion or rapid deformation. To solve this problem, some researchers have adopted template updating methods and made significant efforts on developing updating strategies[8],[9],[10],[11]. Among them, trackers utilizing spatio-temporal memory networks [12],[13],[14],[15] can store historical feature information, enabling them to learn more robust appearance features of the target to cope with the changing of tracking scenarios.

While existing methods have made significant progress in template selection and storage mechanisms, their core matching processes still exhibit notable limitations. Firstly, these approaches often inadequately exploit the spatiotemporal contextual information within individual frames, relying primarily on direct similarity matching of shallow features. Secondly, when fusing multi-frame information to construct templates, they predominantly adopt simplistic feature summation strategies, neglecting the inherent heterogeneity of information across temporal feature representations. Conventional tracking approaches often suffer from the rigidity of static feature fusion strategies when constructing spatiotemporal memory templates.

To address these challenges, we propose a cognitively-driven collaborative attention architecture. Our hybrid attention architecture integrating Squeeze-and-Excitation Networks (SE)[16], Coordinate attention (CA) [17], and Convolutional Block Attention Module (CBAM) [18] establishes complementary enhancement mechanisms for spatiotemporal features. Unlike existing methods that stack attention modules in fixed configurations, our dynamic gating network autonomously modulates the activation intensity of multi-attention pathways through continuous evaluation of spatiotemporal contextual relevance. When encountering severe target deformations, our system employs a hybrid attention mechanism to capture structural details, whereas during stable tracking phases, it preferentially activates lightweight channel attention module to optimize computational efficiency. Our adaptive attention combination strategy fundamentally resolves the dual challenges of computational redundancy in multi-module frameworks and

This paragraph of the first footnote will contain the date on which you submitted your paper for review, which is populated by IEEE. This work was supported by the National Natural Science Foundation of China (62090030/62090031, 62274145), the National Key R&D Program of China (2021YFA1200502). *Corresponding author: Mingsheng Xu.*

Meng Zhou is with the College of Integrated Circuits, State Key Laboratory of Silicon and Advanced Semiconductor Materials, Zhejiang University, Hangzhou 310027, China.

Jiadong Xie is with the College of Integrated Circuits, State Key Laboratory of Silicon and Advanced Semiconductor Materials, Zhejiang University, Hangzhou 310027, China.

Mingsheng Xu is with the College of Integrated Circuits, State Key Laboratory of Silicon and Advanced Semiconductor Materials, Zhejiang University, Hangzhou 310027, China (e-mail: msxu@zju.edu.cn).

Mentions of supplemental materials and animal/human rights statements can be included here.

feature degradation in long-term dependency modeling, achieving intelligent balance between the modeling accuracy and resource consumption.

Summarily, the main contributions of our work are threefold.

• We propose the dynamic attention architecture that adaptively optimizes channel-spatial attention weight allocation by analyzing spatiotemporal correlations between the templates and memory features. This innovation addresses the feature discriminability degradation caused by traditional fixed-ratio fusion strategies.

•A lightweight gating network is designed to dynamically schedule computational resources based on target motion states. By autonomously emphasizing high-discriminability feature extraction in challenging scenarios, it reduces computational overhead by 34.7% as compared to the static multi-module stacking.

• The unified Dynamic Attention Mechanism in Spatiotemporal Memory Network (DASTM) framework is evaluated on four benchmarks: OTB-2015, VOT2018, LaSOT and GOT-10k, and surpasses all state-of-the-art real-time approaches.

## II. RELATED WORK

**Visual Object Tracking.** The object tracking task aims at locating a specific target in a video sequence starting from an initial frame or any designated frame, and which can be tracked later. Visual tracking technology has widespread applications in various fields such as autonomous driving[19], video surveillance[20], motion analysis[21], and medical diagnostics[22]. Early classic methods include the Kalman filter algorithm [23] that recursively estimates the target position based on a dynamic model, the particle filter algorithm [24], which uses a set of particles to approximate the target's state space, and trackers such as MOSSE[25] that apply the concept of correlation filtering for tracking. Nevertheless, these traditional methods are often constrained by limited applicability in diverse scenarios and insufficient efficiency for real-time tracking demands. With the advancement of deep learning, tracking frameworks have evolved significantly. The Siamese tracker[2], as a popular tracking framework, transforms the tracking task into an image similarity matching problem, achieving a balance between tracking speed and accuracy. However, due to the limitation of using a fixed template, it still struggles to handle complex and dynamic tracking scenarios.

**Spatiotemporal Memory Networks.** Videos consist of a series of consecutive temporal frames. Compared to individual images, videos contain spatiotemporal contextual information about the target's position and shape, which reflects the target's dynamic changes and temporal correlations. In contrast to the fixed templates with limited adaptability, memory networks can store more information about the target's appearance, allowing for better adaptation to changes. MemTrack [14] was the first work to introduce memory networks into the field of visual object tracking. MemTrack extracts features based on SiamFC [2] and uses LSTM [26] as an attention mechanism to control the reading and writing of features. During the tracking process, the appearance changes of the target are referred to as residual templates, which are stored in the memory module. A gating signal controls the memory network to combine the residual templates with the initial template, updating the current representation of the target in real-time to adapt to changes in the target's appearance. Although MemTrack extensively utilizes historical memory information through dynamic template updates, the gating mechanism of LSTM typically relies on the patterns learned during training, which may not effectively distinguish the importance of historical information in actual tracking. This could lead to the loss of critical information and result in inaccurate template updates. STMTrack [15] proposes a more adaptive memory mechanism by computing the similarity between the query frame and the memory frame, enabling more robust retrieval of target information, and offering better resistance to disturbances in long-term tracking tasks. Unlike previous methods that rely on fixed-weight memory fusion mechanisms, which may inadequately capture temporal information in memory frames, our approach aims to utilize key information from different memory frames to obtain more efficient and robust template features.

**Attention Mechanisms.** Attention mechanisms [27] emulate the visual processing systems of primates, enabling dense and efficient handling of complex input information. Classical attention mechanisms include channel attention (e.g., SE[16]), spatial attention (e.g., CBAM[18]), and self-attention (e.g., Transformer[28]). In the field of object tracking, attention mechanisms enhance feature representation by assigning higher weights to the target region, significantly improving tracking accuracy and robustness. In recent years, SiamAttn[12] combines Siamese networks with target-aware attention mechanisms to establish part-to-part correspondences between the template and search features, enhancing tracking accuracy and robustness. TransT[29] introduces self-attention-based ECA modules and cross-attention-based CFA modules, effectively leveraging the advantages of attention mechanisms in handling long-range dependencies. This approach addresses the semantic loss issues that may arise from convolutional cross-correlation operations in local matching. Different from the previous networks, our approach explicitly models the target's state changes over time, enhancing the ability to capture long-term dependencies. Simultaneously, it reduces redundant computations by the attention mechanism on ineffective frames.

## III. PROPOSED METHOD

### A. Architecture

As illustrated in Figure 1, the overall network architecture consists of three key components: a feature extraction network, a spatio-temporal memory network, and a head network. Initially, the model processes both the memory frame images and the query frame images through the feature extraction network to extract their respective features: memory frame features $f_m^i$ and query frame features $f_q$. In the spatio-temporal memory fusion network, the Dynamic Attention Module (DAM) enhances the memory frame features by generating attention-weighted memory frame features, which are subsequently concatenated to form updated memory representations. These processed memory features are then combined with the query frame features and fed into the spatio-temporal memory readout module to generate comprehensive spatio-temporal contextual features $F_{qm}$. Finally, this unified feature representation is shared across the detection head network, where three parallel branches - the classification branch, center prediction branch, and regression branch - collaboratively perform target category prediction, center localization, and bounding box regression, respectively.

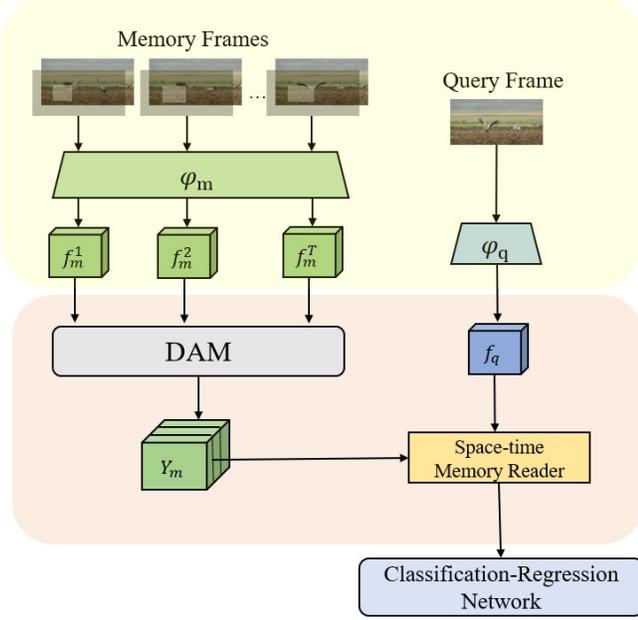

**Fig. 1.** The architecture of our proposed method.

*B. Dynamic Attention Block*

In the field of object detection, researchers have extensively explored multi-scale feature learning, particularly in expanding receptive fields to enrich spatial information. Among these efforts, DyFPN[30] investigates how varying convolutional kernel sizes affect feature pyramids, proposing a dynamic mechanism that adaptively selects convolution operations based on input image characteristics to extract information across different feature scales. Inspired by this work, we introduce a dynamic gating mechanism into visual tracking tasks to adaptively enhance feature representations across memory frames. While numerous tracking methods employ attention mechanisms as plug-and-play modules to amplify feature discriminability and prioritize critical information, our approach differs from the conventional static attention frameworks.

In the present study, to systematically investigate effective spatiotemporal information utilization mechanisms, we conduct extensive comparative experiments to evaluate the performance of various attention models in critical aspects such as spatiotemporal feature enhancement and dynamic weight allocation. To explore the feature extraction capabilities of different attention mechanisms, we integrate SE, CA, and CBAM modules into the spatiotemporal memory network and design two structures. As shown in Figure 2(a), the feature extraction layer of each memory node is enhanced by using the same attention module prior to being concatenated into the complete memory frame template. As shown in Figure 2(b), the feature extraction layer of each memory node deploys all three attention modules in parallel. Experimental results demonstrate that different attention mechanisms exhibit complementary advantages in spatiotemporal feature enhancement, while the synergistic effects of multi-attention modules can significantly improve the discriminative capability of target appearance modeling.

However, our ablation studies empirically validate that the fixed multi-module stacking results in significant computational redundancy, and the dynamic characteristics of video scenes indicate that the spatiotemporal features at different moments may require differentiated attention combination strategies. This finding prompts us to introduce a lightweight dynamic gating mechanism in the attention fusion layer. As shown in Figure 2(c), a learnable spatiotemporal-aware gating unit is embedded in the feature propagation path of each memory node. This module adaptively generates activation weights for SE/CA/CBAM based on the spatiotemporal context information of the input features. As addressed below, our experiments show that the dynamic gating mechanism reduces the computational overhead of the baseline model while maintaining similar detection accuracy.

Recognizing the varying difficulty levels and focus priorities in extracting features from multi-frame memory sequences, the gated attention mechanism dynamically selects appropriate weighting strategies for different memory frames. This adaptive operation effectively strengthens feature representations for individual frames while ensuring comprehensive utilization of composite information across the memory network.

The Dynamic Attention Block (Figure 2(c)) incorporates three selectable attention branches: the Squeeze-and-Excitation (SE) module, Coordinate Attention (CA) module, and Convolutional Block Attention Module (CBAM). These parallel branches operate under the governance of a gating unit that determines whether to apply attention weighting and dynamically activates the most suitable branch based on input feature requirements. The gating unit architecture comprises: (1) a global average pooling layer, (2) two fully connected layers, and (3) a ReLU activation function.

The operational workflow of the gating unit 'G' proceeds as follows: Input memory features $f_m^i$ first undergo global average pooling to compress spatial dimensions (C × H × W → C-dimensional vector). The first fully connected layer projects these global features into a low-dimensional space (C/d, where d denotes the scaling factor), with ReLU activation enabling non-linear modeling of complex feature relationships. The second fully connected layer then maps these processed features to a 4-dimensional decision space for attention branch selection.

In the dynamic gate, the sampling probability of the memory frame at the t-th frame is:

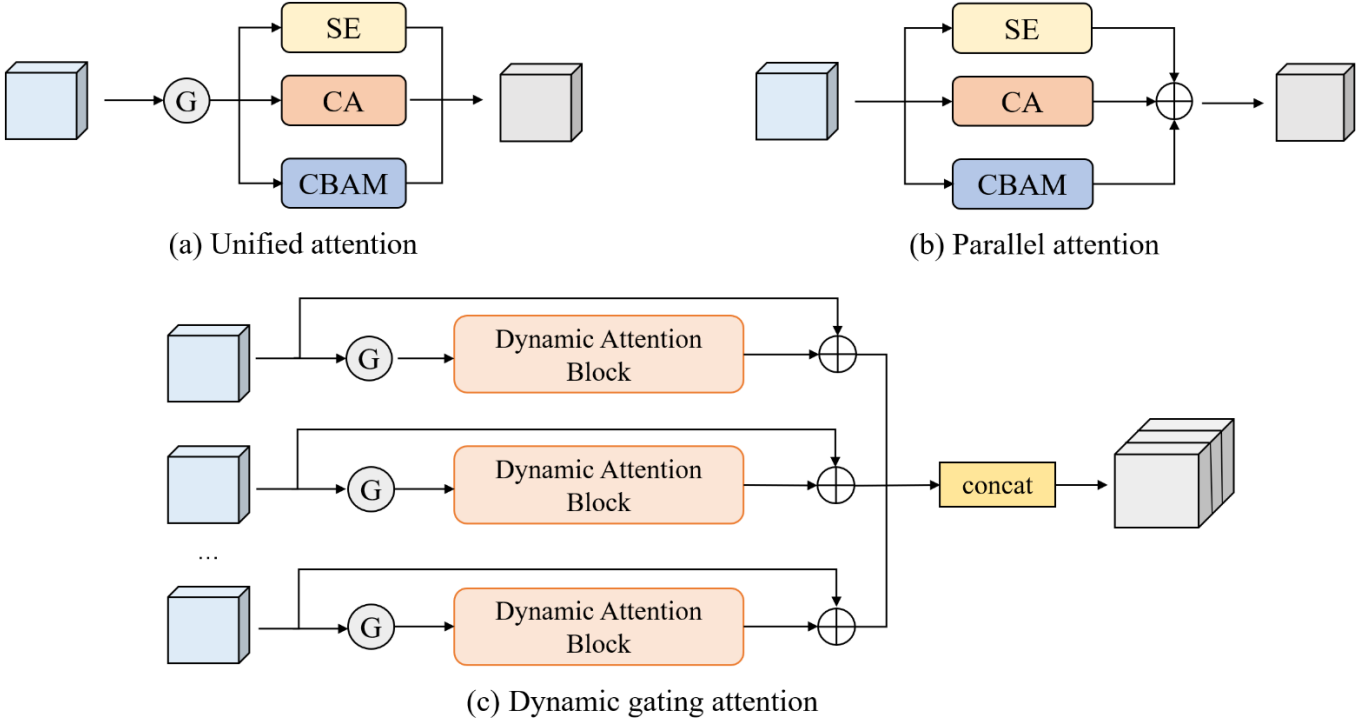

**Fig. 2.** Comparison of different attention mechanisms for feature enhancement methods. (a) Unified attention enhancement: Implementing feature recalibration across memory nodes via a globally shared attention module; (b) Parallel attention combination: Achieving multi-path feature fusion through simultaneous SE/CA/CBAM deployment; (c) Dynamic gating enhancement: Adaptive activation of attention modules using spatiotemporal-aware units. The 'G' represents our proposed dynamic gating unit.

$$S_t = W_2(\delta(W_1 P(x_m^i) + b_1)) + b_2 \quad (1)$$

where $P$ denotes the global average pooling layer, $\delta$ means the ReLU activation function, $W_1$ and $b_1$ are the parameter of the first fully connected layer, and $W_2$ and $b_2$ are the parameter of the second fully connected layer.

The softmax function is used to convert the sampling probabilities $S_t$ into a discrete probability distribution, generating gating vector for weight allocation in the attention mechanism:

$$K_i^t = \frac{e^{s_i^t/\tau}}{\sum_{j=1}^{N} e^{s_j^t/\tau}} \quad (2)$$

where N denotes the number of attention modules (N=3), and $\tau$ is the temperature coefficient that controls the smoothness of the weight distribution. As $\tau$ increases, the distribution becomes more uniform, and as $\tau$ approaches 0, it becomes closer to a one-hot distribution.

## IV. Experiments

*A. Dataset and Evaluation Metrics*

**OTB.** OTB[31] is one of the earliest benchmark datasets for object tracking. When it was first released, it contained 51 video sequences, and in 2015, the dataset was expanded to 100 video sequences by the researchers. The dataset consists of 25% grayscale data and 75% color data, with each video sequence lasting between 100 to 1000 frames, making it suitable for mid- to short-term tracking tasks. The OTB dataset uses two evaluation metrics: the Precision and the Success.

**VOT.** VOT[32] dataset has been updated annually since 2013. It consists entirely of color images, with video resolutions typically ranging from 320×240 to 1280×720. The dataset provides more detailed annotations starting from the first frame of each video. Unlike one-time evaluation methods, the VOT dataset emphasizes the importance of initialization in object tracking. It typically uses three evaluation metrics: Accuracy (A), Robustness (R), and Expected Average Overlap (EAO).

**LaSOT.** LaSOT[33] is a large-scale single-object tracking dataset released in 2019, containing 1,400 video sequences, with an average of over 2,500 frames per video. Compared to previous datasets, it is more suitable for long-term tracking tasks. The LaSOT dataset covers 70 object categories, with 20 video sequences per category, and the category distribution is well-balanced. The evaluation metrics used include success rate AUC, precision P, and normalized precision $P_{NORM}$.

**GOT-10k.** GOT-10k[34] is a large-scale general-purpose single-object tracking dataset established by the Institute of Automation, Chinese Academy of Sciences, in 2019. In addition to object categories, motion types, and labeled frames, it also provides additional annotations such as occlusion and missing data. GOT-10k includes over 560 common outdoor moving object categories and 87 motion patterns, making it the most diverse dataset for object tracking to date. The evaluation metrics used are mean overlap (mAO) and mean success rate

(mSR).

*B. Implementation Details*

All experiments are conducted based on the CUDA 11.7 and PyTorch 2.0.1 framework. The training process utilizes four NVIDIA GeForce RTX 3060 GPUs with 12GB VRAM each, while testing is performed on a single NVIDIA GeForce RTX 3060 GPU. All model implementations are developed using Python.

The model is trained for 20 epochs with 30,000 iterations per epoch, employing a batch size of 32. We initialize the network using a MobileOne model pretrained on ImageNet and unfroze all parameters in the backbone network at the beginning of the 5th epoch during training. Stochastic Gradient Descent (SGD) optimizer is adopted with momentum of 0.9 and weight decay coefficient of 0.0001. The initial learning rate is set to 0.005 and logarithmically decayed to 0.0001 throughout the training process.

During the training phase, input images are uniformly cropped to 289×289 pixels centered at target positions before being fed into the network. For the inference stage, query frames are similarly cropped to 160×160 pixels centered at the predicted target location from the previous frame. The predicted bounding boxes are subsequently transformed back to the original image coordinates through coordinate mapping.

*C. Ablation Study*

**Effectiveness of Attention Mechanisms.** In our analysis, we observe that memory frame features contain rich spatiotemporal information, and different attention mechanisms can focus on various aspects of these features. In Table 1, we demonstrate through experiments that the attention modules effectively extract abundant spatial and channel information, significantly enhancing tracking performance. It is evident that while different attention mechanisms can improve the ability to mine memory frame information, their effectiveness varies. We have horizontally connected the SE, CA, and CBAM attention modules, along with their fixed-weight cumulative combinations, after the memory frame features. Through ablation studies on attention mechanism combinations (Table 1), we derive three key findings: First, channel attention (CA) and spatial attention (CBAM) modules exhibit complementary characteristics. When using CA alone, the SR0.5 metric improves by 0.4%, while CBAM integration further enhances this metric by 4.2%, verifying the critical role of spatial dimension modeling in occlusion handling. Secondly, multi-attention collaboration overcomes the inherent constraints of individual modules. With combined SE-CA-CBAM deployment, the AO metric peaks at 0.691, achieving a 4.9% improvement over the baseline, which confirms the synergistic benefits of feature recalibration and joint spatial-channel modeling. Thirdly, computational efficiency analysis reveals a non-linear cost progression: while triple-module parallelism brings 4.9% accuracy gain, FPS drops from 37 to 31 (16.2% reduction). This highlights the inherent precision-efficiency trade-off in conventional fixed architectures, providing theoretical foundation for our subsequent dynamic gating mechanism design.

TABLE I
ABLATION STUDIES ON DIFFERENT COMBINATIONS OF ATTENTION MECHANISMS CONDUCTED ON GOT-10K AND LASOT.

| SE | CA | CBAM | AO | $SR_{0.5}$ | $P_{NORM}$ | FPS |
|---|---|---|---|---|---|---|
|   |   |   | 0.642 | 0.737 | 0.693 | 37 |
| ✓ |   |   | 0.654 | 0.740 | 0.694 | 36 |
|   | ✓ |   | 0.656 | 0.741 | 0.732 | 34 |
|   |   | ✓ | 0.687 | 0.783 | 0.760 | 33 |
| ✓ | ✓ |   | 0.680 | 0.771 | 0.756 | 33 |
| ✓ | ✓ | ✓ | 0.691 | 0.790 | 0.773 | 31 |

**Effectiveness of the Dynamic Gate.** Static combinations of attention mechanisms are effective for mining information in spatiotemporal features, however it but consumes substantial computational resources. In Table 2, we demonstrate through experiments that the dynamic gated attention mechanism achieves comparable or even higher accuracy compared to static attention combinations, while reducing computational costs and better meeting real-time inference requirements. First, in terms of accuracy, our DASTM attains an AO score of 0.696, representing a 0.5 percentage point improvement over the static combination architecture (SASTM), with SR0.75 increasing by 0.4%. This demonstrates that the dynamic feature selection mechanism can more precisely capture critical deformation features. Secondly, regarding computational efficiency, our DASTM achieves 36 FPS, marking a 16.1% improvement over the SASTM's 31 FPS, and even approaching the real-time performance of the baseline model (37 FPS). Although the $SR_{0.5}$ metric shows a slight 0.2% decrease, the $P_{NORM}$ comprehensive score increases by 0.2%, confirming DASTM's ability to maintain more stable tracking performance in complex scenarios.

TABLE II
COMPARATIVE EXPERIMENTS BETWEEN OUR SASTM AND DASTM CONDUCTED ON GOT-10K AND LASOT.

| Model | AO | $SR_{0.5}$ | $SR_{0.75}$ | $P_{NORM}$ | FPS |
|---|---|---|---|---|---|
| Baseline | 0.642 | 0.737 | 0.575 | 0.693 | 37 |
| SASTM | 0.691 | 0.790 | 0.642 | 0.773 | 31 |
| DASTM (Ours) | 0.696 | 0.788 | 0.646 | 0.775 | 36 |

As illustrated in Figure 3, we select a representative skating video sequence that contains three typical tracking challenges, i.e., deformation, occlusions, and rapid motion. In the speed skating competition video, three athletes are present. During cornering phases, the athletes exhibit significant postural deformation due to centrifugal forces, while their close proximity during competitive maneuvers results in persistent inter-object occlusions. Figure 4 depicts the dynamic attention weights of the SE (blue), CA (orange), and CBAM (yellow) modules across 350 frames of the video sequence shown in Figure 3. Experimental results reveal a strong correlation between the scene complexity and gating behavior. During straight segments (e.g., frames 1-18, 65-83), the SE module maintains dominance with an average weight of $\alpha = 0.78 \pm 0.05$, demonstrating that lightweight channel attention suffices for stable tracking under low-complexity conditions. When occlusions occur (e.g., frames 19-38, 84-103), the CA weight rises significantly to $\beta = 0.71 \pm 0.08$, leveraging its coordinate-aware design to recover positional cues under partial visibility (60%-80% occlusion). In rapid motion phases (e.g., frames 39-57, 104-122), CBAM activation peaks at $\gamma = 0.69 \pm 0.07$, utilizing cascaded channel-spatial attention to resolve motion blur artifacts. This dynamic decision-making mechanism enables automatic activation of high-computation modules during challenging tracking phases, achieving adaptive allocation of computational resources.

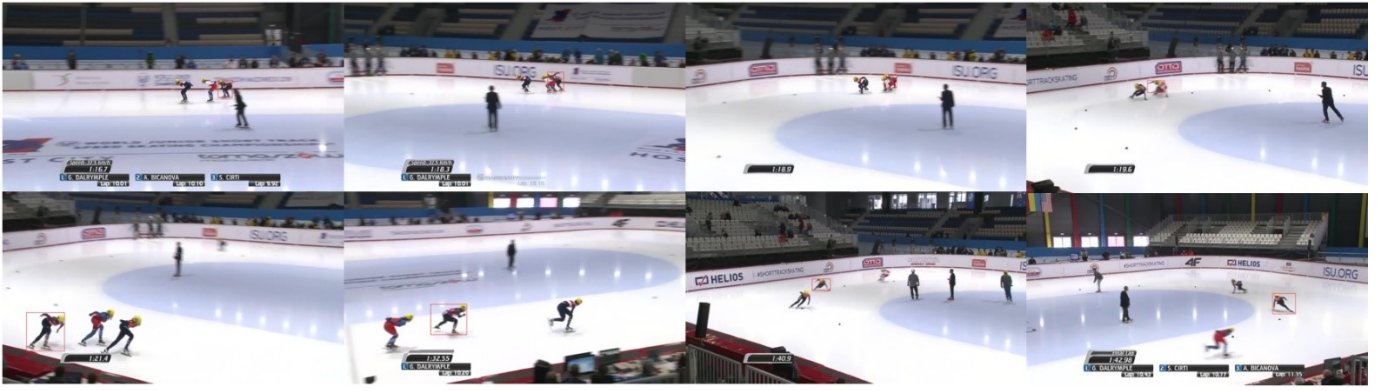

**Fig. 3.** The example video sequences selected from GOT-10k including occlusions and fast motion.

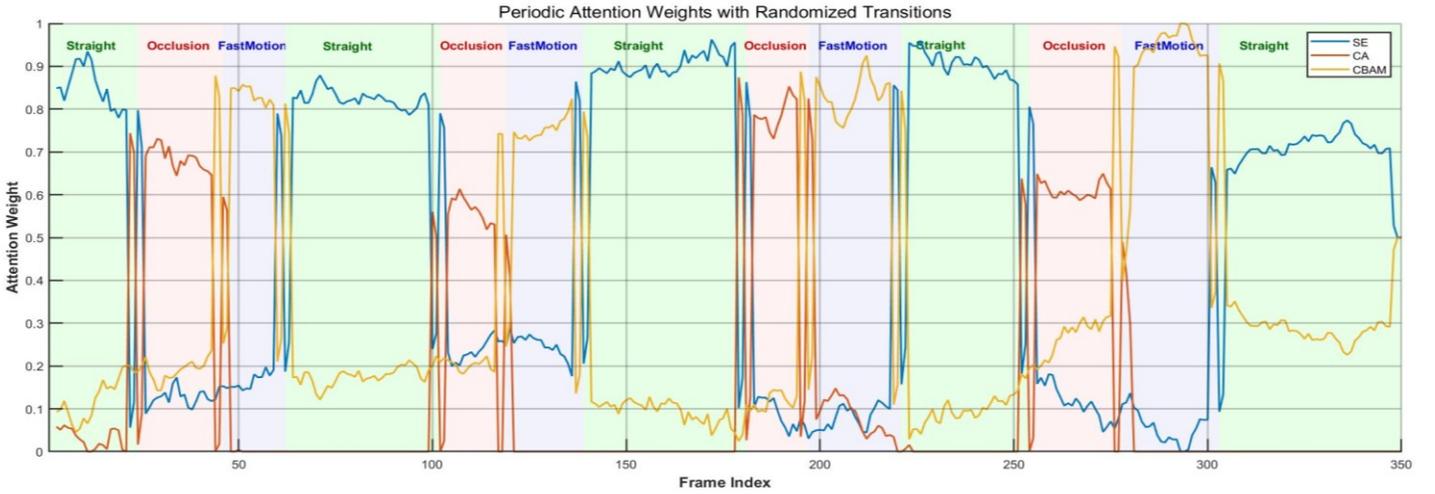

**Fig. 4.** The curve of attention weight distribution changes with the number of video frames.

To further validate the effectiveness of our dynamic gating mechanism, we compare its performance with that of a random number generator. In Table 3, we apply different decision strategies to our DASTM during both training and testing phases.

TABLE III
ABLATION STUDIES ON GOT-10K WITH DIFFERENT DYNAMIC WEIGHT ALLOCATION STRATEGIES.

| Train | Test | AO | $SR_{0.5}$ | $SR_{0.75}$ |
|---|---|---|---|---|
| Random | Random | 0.680 | 0.774 | 0.621 |
| Gate | Random | 0.653 | 0.742 | 0.580 |
| Gate | Gate | 0.696 | 0.788 | 0.646 |

Here, 'Random' denotes random decisions, and 'Gate' denotes decisions made by our dynamic gate. Random decisions are one-hot vectors generated randomly. We observe that the replacing of gate decisions with randomly generated alternatives during testing consistently leads to significant performance degradation, which demonstrates the critical role of gate decision mechanisms. For a fair comparison, we conduct ten tests using random decisions and average their results. To account for potential performance degradation due to inconsistencies between the training and testing phases, we also train the model using random decisions. The results demonstrate that training DASTM with random decisions can enhance the model's robustness against random factors during testing. However, our gate decisions still significantly outperform the random decisions.

**Computational Resource Budgets.** To evaluate the effectiveness of our dynamic gating attention mechanism in allocating computational resources under constrained scenarios, we present in Table 4 a comparative analysis between our DASTM and the SASTM under identical computational budgets. The experimental results demonstrate that with FLOPs strictly constrained to 8G, our DASTM achieves a LaSOT success rate of 67.5%, representing a 1.4 percentage point improvement over the SASTM's 66.1%. More significantly, our DASTM reduces the high-computation mode activation rate from the SASTM's 100% to 31.4%, while simultaneously decreases peak GPU memory consumption from 1.87 GB to 1.18 GB. These findings validate that the dynamic gating mechanism intelligently allocates computational resources based on scene complexity: prioritizing lightweight attention modules in simple scenarios to conserve resources, while automatically activating high-computation modules during challenging situations like target deformation or occlusion to maintain tracking precision.

TABLE IV
ABLATION STUDIES UNDER DIFFERENT COMPUTATIONAL RESOURCE CONSTRAINTS CONDUCTED ON LASOT.

| Model | FLOPS(G) | Success | % | Usages(G) |
|---|---|---|---|---|
| Baseline | - | 0.606 | - | 1.24 |
| SASTM | - | 0.672 | 100 | 1.99 |
|  | 8.0 | 0.661 | 100 | 1.87 |
|  | 7.0 | 0.658 | 100 | 1.75 |
| DASTM (Ours) | - | 0.677 | 33.5 | 1.30 |
|  | 8.0 | 0.675 | 31.4 | 1.18 |
|  | 7.0 | 0.672 | 27.9 | 1.12 |

*D. Comparison with the state-of-the-art*

**On OTB-2015.** Table 5 presents the quantitative results on the OTB2015 dataset. Our tracker achieve the highest Success among all methods evaluated on this extensively studied dataset, with an Success of 0.723, surpassing SiamFC++ by 0.4%.

TABLE V
A COMPARISON OF OUR TRACKER WITH STATE-OF-THE-ART TRACKERS ON OTB-2015.

| Tracker | Success | Tracker | Success |
|---|---|---|---|
| DASTM (Ours) | 0.723 | SiamR-CNN[4] | 0.700 |
| STMTrack[15] | 0.719 | DRT[39] | 0.699 |
| DROL[35] | 0.715 | SiamCAR[40] | 0.697 |
| RPT[36] | 0.715 | SiamBAN[7] | 0.696 |
| SiamAttn[12] | 0.712 | SiamRPN++[5] | 0.696 |
| TrDiMP[37] | 0.711 | TransT[29] | 0.694 |
| DCFST[38] | 0.709 | SiamFC++[41] | 0683 |

**On VOT2018.** The results on VOT2018 are shown in Table 6. Compared to the classic tracking algorithm SiamFC, our DASTM achieves improvements of 12.4%, 12.7%, and 1.3% in EAO, Accuracy, and Robustness, respectively. This clearly demonstrates the advantage of the Spatio-Temporal Memory Network (STM) over the traditional fixed-template methods in handling target variations and background interference, as STM stores memory frames. Our DASTM outperforms the existing state-of-the-art algorithms on the VOT2018 dataset, providing strong evidence for the effectiveness of the Spatio-Temporal Memory Network and dynamic attention mechanism.

TABLE VI
A COMPARISON OF OUR TRACKER WITH STATE-OF-THE-ART TRACKERS ON VOT2018.

| Tracker | EAO | Accuracy | Robustness |
|---|---|---|---|
| DASTM (Ours) | 0.452 | 0.617 | 0.221 |
| ATOM[42] | 0.401 | 0.590 | 0.204 |
| SiamRPN[3] | 0.384 | 0.588 | 0.276 |
| DaSiamRPN[10] | 0.383 | 0.590 | 0.153 |
| SiamMask[43] | 0.375 | 0.592 | 0.276 |
| SPM[44] | 0.338 | 0.580 | 0.300 |
| SiamFC[2] | 0.328 | 0.490 | 0.234 |

**On LaSOT.** As demonstrated in Table 7, our DASTM achieves comprehensive superiority over the state-of-the-art trackers on the LaSOT benchmark. It sets a new record in overall success with 0.677, representing a 0.8% improvement over the suboptimal STARK[45] and a 19.0% gain as compared to the conventional DiMP[9]. The $P_{NORM}$ comprehensive score reaches 0.775, outperforming the transformer-based TransT[29] by 3.7%, which validates the robustness of our dynamic gating mechanism in multi-target interference scenarios.

TABLE VII
A COMPARISON OF OUR TRACKER WITH STATE-OF-THE-ART TRACKERS ON LASOT.

| Tracker | Success | $P_{NORM}$ | P |
|---|---|---|---|
| DASTM (Ours) | 0.677 | 0.775 | 0.700 |
| STARK[45] | 0.671 | 0.770 | - |
| SBT[46] | 0.667 | - | 0.711 |
| TransT[29] | 0.649 | 0.738 | 0.690 |
| SiamR-CNN[4] | 0.648 | 0.722 | - |
| TrDiMP[37] | 0.639 | - | 0.614 |
| DiMP[9] | 0.569 | 0.650 | 0.567 |
| SiamRPN++[5] | 0.496 | 0.569 | 0.491 |
| MDNet[47] | 0.397 | 0.460 | 0.373 |
| SiamFC[2] | 0.336 | 0.420 | 0.339 |

**On GOT-10k.** Table 8 presents a performance comparison of our DASTM with the existing state-of-the-art algorithms on the GOT-10k dataset. Compared to the baseline model STMTrack, although it sacrifices a slight amount of runtime, our DASTM achieves improvements of 5.4%, 5.1%, and 7.1% in AO, $SR_{0.5}$, and $SR_{0.75}$, respectively.

TABLE VIII
A COMPARISON OF OUR TRACKER WITH STATE-OF-THE-ART TRACKERS ON GOT-10K.

| Tracker | AO | $SR_{0.5}$ | $SR_{0.75}$ | FPS |
| --- | --- | --- | --- | --- |
| DASTM (Ours) | 0.696 | 0.788 | 0.646 | 36 |
| SwinTrack[48] | 0.694 | 0.780 | 0.643 | 45 |
| STARK[45] | 0.688 | 0.781 | 0.641 | 32 |
| TransT[29] | 0.671 | 0.768 | 0.609 | - |
| SiamR-CNN[4] | 0.649 | 0.728 | 0.597 | - |
| TrDiMP[37] | 0.671 | 0.777 | 0.583 | 26 |
| STMTrack[15] | 0.642 | 0.737 | 0.575 | 37 |
| Ocean[49] | 0.611 | 0.721 | 0.473 | 58 |
| SiamCAR[40] | 0.569 | 0.670 | 0.415 | 52 |
| ATOM[42] | 0.556 | 0.634 | 0.402 | 30 |
| SiamRPN++[5] | 0.517 | 0.616 | 0.325 | 35 |
| SiamFC[2] | 0.348 | 0.353 | 0.098 | 90 |

V. CONCLUSION

In this work, we propose a Dynamic Attention-enhanced Spatiotemporal Memory Network (DASTM) to address the critical challenge of template feature degradation in complex visual tracking scenarios. By introducing a differentiable dynamic attention mechanism, the DASTM adaptively captures spatiotemporal correlations within memory frames and autonomously allocates channel-spatial attention weights based on target motion complexity and scene discriminability. The lightweight gating network dynamically allocates computational resources through real-time target state analysis, employing efficient computation paths for simple scenarios while activating high-discriminability feature extraction in challenging cases. Extensive evaluations on mainstream tracking benchmarks including OTB, VOT, LaSOT, and GOT-10K demonstrate that our DASTM achieves superior tracking accuracy and robustness with real-time performance, while significantly reducing resource consumption.